\pdfoutput=1

\documentclass[11pt]{article}

\usepackage[]{acl}

\usepackage{times}
\usepackage{latexsym}

\usepackage[T1]{fontenc}

\usepackage[utf8]{inputenc}

\usepackage{microtype}

\usepackage{amsmath}
\usepackage{enumitem}
\usepackage{float}
\usepackage{graphicx}
\usepackage{multirow}

\usepackage[linesnumbered, ruled]{algorithm2e}
\usepackage[T1]{fontenc}
\usepackage{dsfont}

\usepackage{mdframed}
\usepackage{booktabs}
\usepackage{graphicx}
\usepackage{subcaption}

\usepackage[normalem]{ulem}
\usepackage{enumitem}
\usepackage{xcolor}
\usepackage{soul}
\colorlet{soulred}{red!30}
\sethlcolor{soulred}%

\usepackage{amsmath,amsthm,amsfonts,amssymb,bm}
\usepackage{framed}
\usepackage{varioref}
\usepackage{float}
\usepackage{multirow}
\usepackage{dashrule}
\usepackage{url}
\usepackage{pifont}
\usepackage{helvet}
\usepackage{colortbl}
\usepackage{xcolor}
\usepackage{multirow}
\usepackage{booktabs}

\definecolor{lightgray}{gray}{0.9}
\newcommand{\ie}{\emph{i.e., }}
\newcommand{\eg}{\emph{e.g., }}
\newcommand{\etal}{\emph{et al. }}

\definecolor{green}{rgb}{0.1,0.1,0.1}
\definecolor{chocolate}{HTML}{D2691E}
\definecolor{maroon}{HTML}{A00000}
\definecolor{indigo}{HTML}{4B0082}
\definecolor{green}{HTML}{008000}
\definecolor{cadmiumgreen}{rgb}{0.0, 0.42, 0.24}

\usepackage{amssymb}
\usepackage{pifont}

\makeatletter
\newcommand*\myfontsize{%
  \@setfontsize\myfontsize{8}{9}%
}
\makeatother

\makeatletter
\newcommand*\mysmallfontsize{%
  \@setfontsize\mysmallfontsize{7.4}{8.4}%
}
\makeatother

\usepackage{listings}
\usepackage{xcolor}
\usepackage{adjustbox}

\newcommand{\myskip}[1]{}

\title{FIHA: Automated Fine-grained Hallucinations Evaluations in \\ Large Vision Language Models with Davidson Scene Graphs}

\author{
Bowen Yan\thanks{\hspace{1em}Equal Contribution} \quad Zhengsong Zhang\footnotemark[1] \quad Liqiang Jing\footnotemark[1] \quad Eftekhar Hossain \quad Xinya Du\thanks{\hspace{1em}Corresponding Author} \\
University of Texas at Dallas, Richardson, United States \\
\texttt{\{bowen.yan, zhengsong.zhang, liqiang.jing, xinya.du\}@utdallas.edu}
}

\begin{document}
\maketitle

\begin{abstract}

The rapid development of Large Vision-Language Models (LVLMs) often comes with widespread hallucination issues, making cost-effective and comprehensive assessments increasingly vital. 
Current approaches mainly rely on costly annotations and are not comprehensive -- in terms of evaluating all aspects, such as relations, attributes, and dependencies between aspects. 
Therefore, we introduce the FIHA (autonomous \underline{F}ine-gra\underline{I}ned \underline{H}allucination ev\underline{A}luation in LVLMs), which could access LVLMs hallucination in an LLM-free and annotation-free way and model the dependency between different types of hallucinations. 
FIHA can generate Q\&A pairs on any image dataset at minimal cost, enabling hallucination assessment from both image and caption. Based on this approach, we introduce a benchmark called FIHA-v1, which consists of diverse questions on various images from three datasets. Furthermore, we use the Davidson Scene Graph (DSG) to organize the structure among Q\&A pairs, in which we can increase the reliability of the evaluation. We evaluate representative models using FIHA-v1, highlighting their limitations and challenges. We released our code and data at \url{https://github.com/confidentzzzs/FIHA}.


\end{abstract}

\section{Introduction}
\label{Method}

\begin{table*}
  \caption{Comparison with other benchmarks. Dis. denotes Discriminative and Gen. denotes Generative.}
  \label{sample-table}
  \centering
  \small
  \resizebox{\textwidth}{!}{
    \begin{tabular}{lccccccccc}
      \toprule
      \multirow{3}{*}{Evaluation Methods} &  \multicolumn{3}{c}{Discriminative Hallucination} & \multicolumn{2}{c}{Task Type} & \multirow{3}{*}{Dependency} & \multirow{3}{*}{LLM Free}& \multirow{3}{*}{Annotation Free}\\
      \cmidrule(r){2-4} 
      \cmidrule(r){5-6}
      & Object & Attribute & Relation & Dis. & Gen. & & & \\
      \midrule
      
      POPE \cite{li2023evaluating} & \checkmark & $\times$ & $\times$ & \checkmark & $\times$ & $\times$ & \checkmark & \checkmark\\
      NOPE \cite{lovenia2023negative} & \checkmark & $\times$ & $\times$ & \checkmark & $\times$ & $\times$ & $\times$ & \checkmark\\
      CIEM \cite{hu2023ciem} & \checkmark & \checkmark & $\times$ & \checkmark & $\times$ & $\times$ & $\times$ & \checkmark\\
      Bingo \cite{cui2023holistic}  & $\times$ & $\times$ & $\times$ & $\times$ & \checkmark & $\times$ & \checkmark &$\times$ \\
      AMBER \cite{wang2023llm} & \checkmark & \checkmark & \checkmark & \checkmark & \checkmark & $\times$ & \checkmark & $\times$\\
      HallusionBench  \cite{liu2023hallusionbench} & $\times$ & $\times$ &$\times$  & $\times$ & \checkmark & $\times$ & \checkmark &$\times$ \\
      MHaluBench \cite{chen2024unified} & \checkmark & \checkmark & $\times$ & $\times$ & \checkmark & $\times$ & $\times$ & $\times$\\
      Hal-Eavl \cite{jiang2024hallucination} &\checkmark  & \checkmark &\checkmark  &\checkmark  & \checkmark & $\times$ & $\times$ &\checkmark \\
      FIHA (ours) & \checkmark & \checkmark & \checkmark & \checkmark & \checkmark & \checkmark & \checkmark & \checkmark \\
      \bottomrule
    \end{tabular}
  }
\end{table*}

Large Vision-Language Models (LVLMs) such as MiniGPT-4 ~\cite{minigpt4} and LLaVA ~\cite{liu2023llava}, which extend Large Language Models (LLMs) by incorporating visual encoders, such as CLIP \cite{DBLP:conf/icml/RadfordKHRGASAM21}, have shown prominent capabilities in visual understanding and generation~\cite{zhang2024vision}. 
However, LVLMs suffer from the issue of hallucination, which can lead to misinterpretation or erroneous assertions of the visual inputs, thus hindering the performance of models in multi-modal tasks~\cite{huang2023survey,DBLP:journals/corr/abs-2402-11414,DBLP:conf/aaai/ZhangJG25}.
Specifically, the models may describe objects that do not exist in the image or incorrect object attributes and relations between objects. Generating such unreliable content will greatly reduce the model's credibility. Therefore, it is crucial to establish a benchmark for evaluating the hallucination level of LVLMs. 
\par

Previous studies \cite{li2023evaluating,wang2023evaluation,wang2023llm}, as shown in Table \ref{sample-table}, primarily employ a Question Generation (QG) module to create a set of validation questions and expected answers (\ie Q\&A pairs) for hallucination evaluation. These generated questions are then used to evaluate hallucinations in LVLMs.
Despite the compelling success of the existing work, they still face two main challenges: (1) The existing work overlooks the dependency between different kinds of questions. For example, if the answer to ``Is there a bike?'' is no, dependent questions like ``Is the bike yellow?'' should be skipped, detailed explanations can be found in the Appendix \ref{app:explanation}. 
(2) Additionally, most prior work heavily relies on human annotations \cite{wang2023llm} or LLMs \cite{li-etal-2023-halueval} to generate Q\&A pairs used in hallucination evaluation, which can be costly or labor-intensive.

To mitigate these limitations, we propose Fine-grained Hallucination Evaluation (FIHA), an automatic evaluation framework for assessing fine-grained and diverse hallucinations in large-scale vision-language models. The framework accepts either images or captions as input and generates Q\&A pairs by extracting objects, attributes, and entity relations. It then formulates diverse questions (e.g., ``what'', ``who'', ``which'', etc.) that allow for free-form responses. By integrating BLIP-2 \cite{li2023blip} for caption generation, Fast R-CNN \cite{girshick2015fast} for feature extraction, and a question-generation template, our pipeline enables fully automatic Q\&A generation without relying on LLMs or manual annotations.

We organize Q\&A pairs using the Davidson Scene Graph (DSG)~\cite{cho2023davidsonian}. The DSG ensures leaf node responses depend on root node answers, reducing errors and improving reliability. Our Q\&A pairs include negative, narrative, and interrogative questions, allowing a progressive, comprehensive evaluation of image understanding.

\par We make the following key contributions through this work:
\begin{itemize}[leftmargin=*]
    \item To the best of our knowledge, FIHA is the first automated hallucination evaluation framework that is LLM-free and annotation-free. This approach not only scales efficiently but also minimizes labor and associated costs.
    \item Based on FIHA, we generate a DSG-based fine-grained evaluation benchmark FIHA-v1 that includes Q\&A pairs evaluating various types of hallucinations and the semantic dependency relation organized by DSG.
    \item We evaluate and analyze several mainstream open-source and close-source LVLMs with FIHA-v1, providing valuable insights into their performance.
\end{itemize}

\section{Methodology}
\label{sec:method}

\begin{figure*}[h]
  \centering
  \includegraphics[width =1\linewidth]{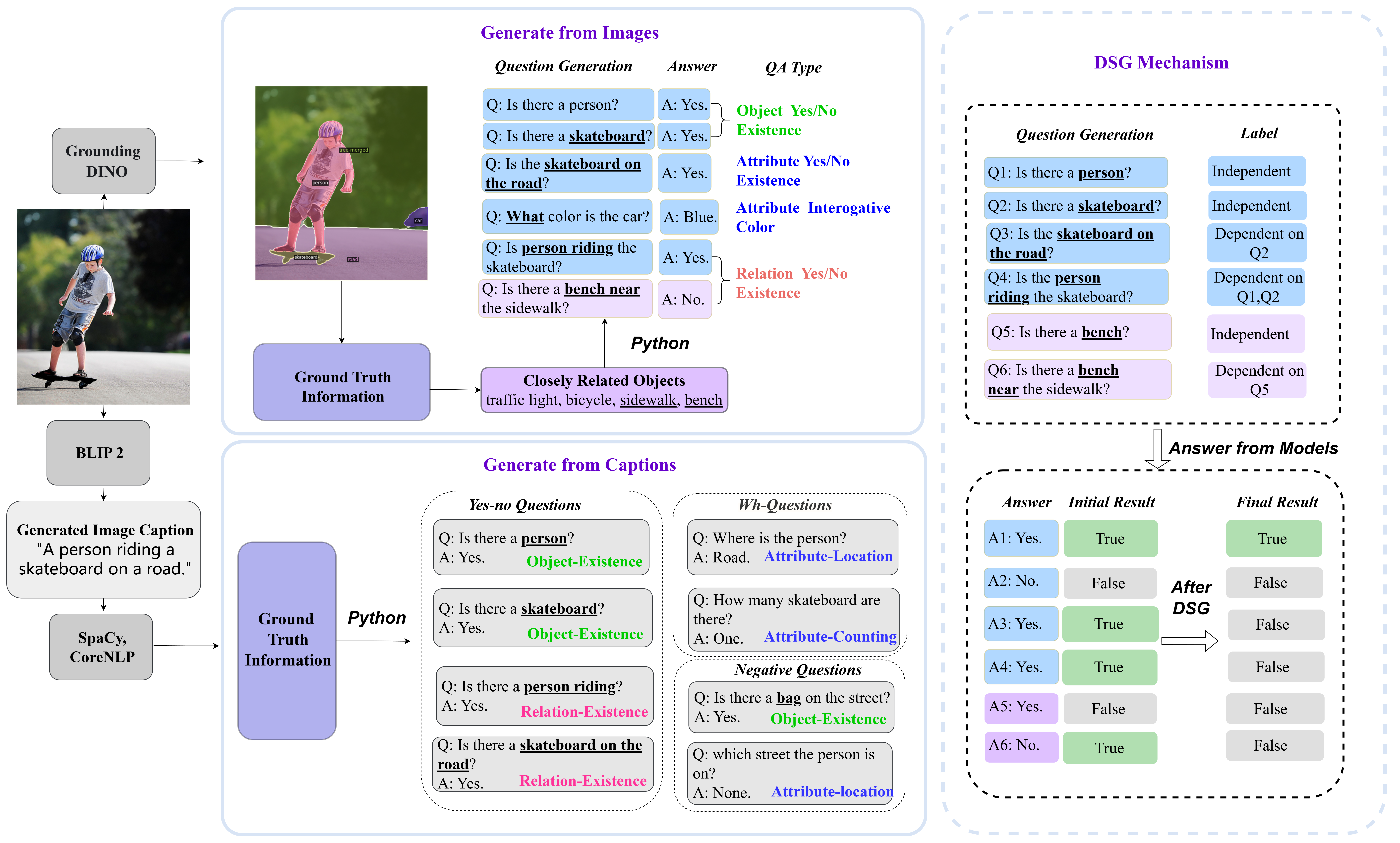}
  \vspace{-1mm}
 \caption{Overview of FIHA framework. FIHA extracts entities, attributes, and relations from images and captions respectively, and generates comprehensive and diverse questions to thoroughly detect hallucinations. 
In the Figure, we can see that no large language model (LLM) \cite{achiam2023gpt} or additional manual annotations are used.} 
\vspace{-3mm}
 \label{block}
\end{figure*}
In this section, we introduce the overall pipeline of FIHA as illustrated in Figure \ref{block}. In summary, our pipeline offers two approaches for Q\&A generation. The first is based on images: given an image \( I \), we extract the necessary entities, including features such as objects, object attributes, and relations. Using a rule-based method, we then generate the Q\&A pairs. The second approach is based on captions. If an image does not already have a caption, we can use BLIP-2 \cite{li2023blip} to generate captions. Alternatively, if the dataset includes original captions, we can use them directly as input, pass them through the feature extraction process, and generate the corresponding Q\&A pairs.

\par

\subsection{Fine-grained Information Extraction}

\subsubsection{Information Acquisition from the Caption}
Q\&A generation based on caption includes caption generation (optional if original datasets include captions) and extract information (object existence, object attributes and object relations) and using these information to generate Q\&A pairs.

\textit{Caption Generation}.
Image captions can depict an image in detail, demonstrating fine-grained visual information, such as objects, attributes and relations. 
Inspired by the findings of previous research \cite{li2023evaluating}, which indicate that smaller vision-language models tend to produce more concise responses with fewer hallucinations compared to mainstream LVLMs. As such, we select BLIP-2 to generate a caption for the image. This step allows us to generate highly credible captions based on the image.
\par
\textit{Fine-grained Information Extraction.} 
In this process, we take either the generated captions (if the ground-truth caption is not available) or the ground-truth captions, depending on the user's needs, as input and extract information such as object existence, object attributes, and relations from the captions. For extracting objects and attributes in the caption, we use SpaCy's \cite{spacy2} part-of-speech tagging feature to identify objects and their corresponding attributes, such as numerals, adjectives, and verbs. 
As a result, we obtain all the ground truth objects and their attributes as: $G^C_{O,A} = \{ o_1: A_1, o_2 : A_2, \dots, o_n: A_n \}
$, where $n$ is the number of objects. 
$o_i$ is the $i$-th object and $A_i$ is all attributes for the $i$-th object.
Relations from the captions are extracted using the Stanford CoreNLP library\footnote{\url{https://stanfordnlp.github.io/CoreNLP/}.}, which provides a powerful suite of NLP tools for performing various linguistic analyses on text, making it an ideal choice for relation extraction. From this process, we obtain all the relations: $G^C_R = \{ R_1 (o_{R_1}^1,o_{R_1}^2), \dots, R_m((o_{R_m}^1,o_{R_m}^2))\}$, where $m$ is the number of relations. 
$R_i$ is the $i$-th relation for the objects $o_{R_i}^2$ and $o_{R_i}^2$.



\subsubsection{Information Acquisition from the Image}
As the caption may lose some information in the image, our second approach to extract information is directly from images, which provides richer and more detailed information than captions alone. 
For object and attribute extraction, we use Grounding DINO \cite{liu2023grounding}, a well-established and widely used object detection method based on Transformer architecture. 
Grounding DINO has been a pioneering approach in the field of object detection due to its ability to quickly identify objects within an image while simultaneously predicting their attributes. 
This method allows us to retrieve the ground truth objects along with their corresponding attributes such as color, size, and shape, forming a set of objects and attributes: 
$G^I_{O,A} = \{ o_1: A_1, o_2 : A_2, \dots, o_n: A_n \}$, where $n$ represents the number of objects detected. 
In addition to identifying objects and their attributes, it is crucial to understand how these objects interact or relate within the scene. 
For this purpose, we employ RelTR \cite{2022RelTR}, a cutting-edge method designed to generate sparse scene graphs by decoding visual appearances and learning both subject and object queries from the image data. RelTR enables us to extract meaningful relationships between the detected objects, such as spatial relations (e.g., one object being behind or near another) and actions (e.g., wearing or holding), resulting in a set of ground truth relations: $G^I_R = \{ R_1 (o_{R_1}^1,o_{R_1}^2), \dots, R_m((o_{R_m}^1,o_{R_m}^2))\}$, where $m$ denotes the number of relationships extracted from the image. By combining both object detection and relational extraction, this approach provides a comprehensive understanding of the visual content, which is essential for generating accurate and meaningful Q\&A pairs.

\begin{figure}[h]
\label{extracted information}
  \centering
  \includegraphics[width=1\linewidth]{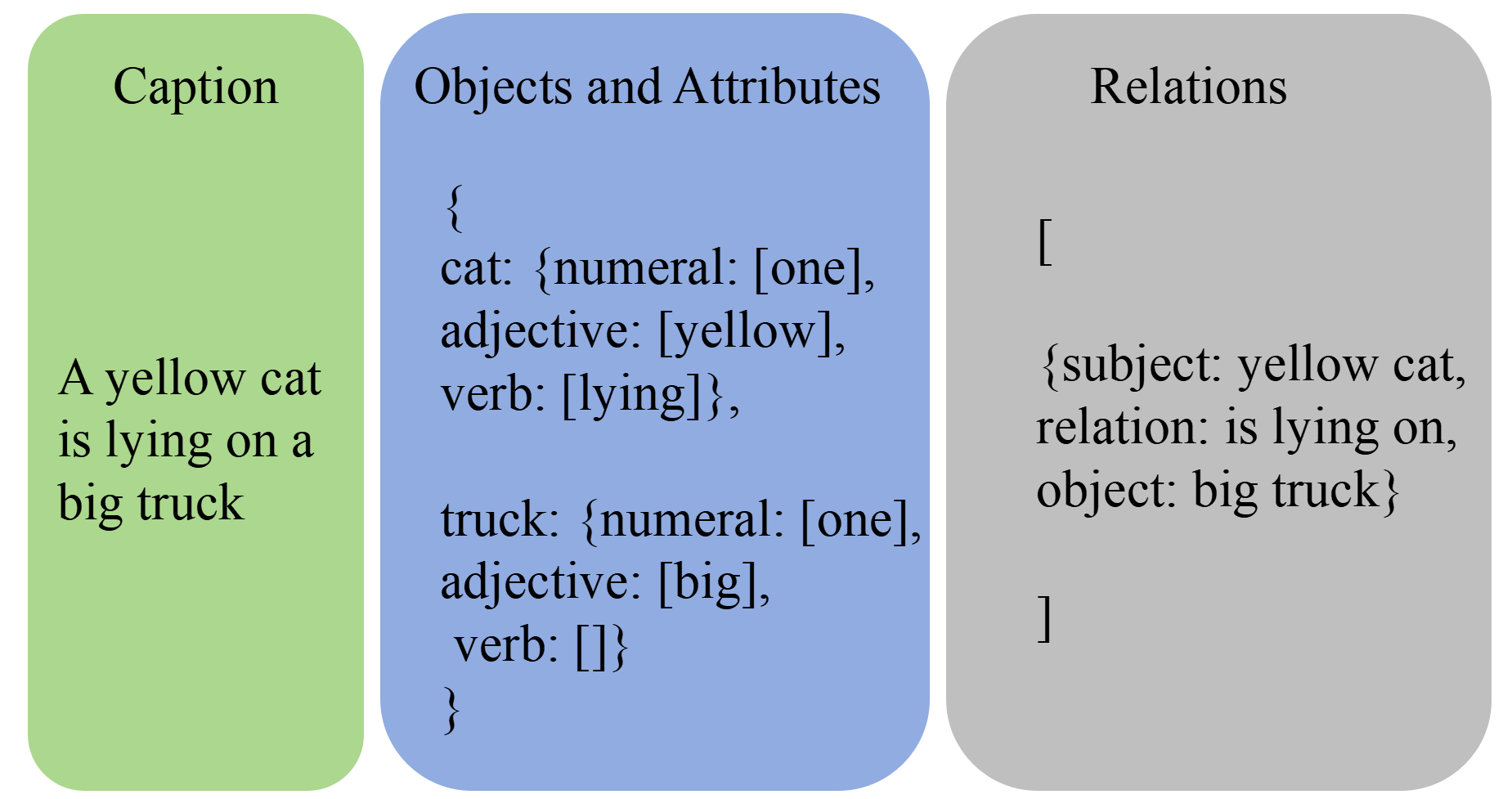}
  \caption{Example of extracted information.}
  \label{caption_obj_att_rel}
\end{figure}

\subsection{Question Answer Pair Generation}
\label{Question Answer Pair Generation}

Next, we generate corresponding Q\&A pairs for the extracted image and text information, respectively. After reorganizing them through DSG, they can be directly input into the model to detect the level of hallucination.

We use two kinds of questions for hallucination evaluation. The first type is Yes-No questions, which check object existence, such as ``Is there any \{obj$_k$\}?" and object relations like ``Is there a \{obj$_1$\} near the \{obj$_k$\}?". These questions help determine whether specific objects and their relationships are present within an image. Additionally, Yes-No questions assess attributes by asking about features like color, size, or location.

The second type is Wh-Questions, which add diversity to the evaluation by incorporating interrogative words such as ``what'', ``who", ``which'', ``where'', and ``how many''. These questions elicit more detailed, free-form responses, typically no longer than three words. For example, ``What color is the \{obj$_k$\}?" or ``Which object is near the \{obj$_k$\}?" help assess the finer details about objects and their relationships. Unlike traditional hallucination evaluations that primarily rely on Yes-No questions, our approach includes both types to provide a more comprehensive assessment. 

We introduce Negative Questions for both Yes-No and Wh-Questions. These questions are created by replacing real objects, attributes (e.g., color, size), and relations in the original Q\&A pairs with non-existent ones. For example, "Is there a car here?" becomes "Is there a plane here?" using a randomly selected object from a pre-constructed set. To ensure accurate evaluation, we avoid including objects present in the current image in the candidate set. These questions are answered with negative pronouns like "none," "nobody," or "nowhere."

\subsection{Davidson Scene Graph}
\label{sec:dsg}

To model the dependency between objects, attributions, and relations accurately and improve the reliability of hallucination evaluation,
we introduce the Davidsonian Scene Graph (DSG) \cite{cho2023davidsonian} mechanism. The DSG schematic diagram can refer to Figures \ref{block} and \ref{dsg_figure}. The DSG can be understood as a post-processing step for the Q\&A pairs. After obtaining all the Q\&A pairs, we organize them into multiple tree-like structures, where each Q\&A pair serves as a node. 
According to the structure of the tree-like structures, each node is either a root node or a leaf node. Specifically, the entire process is divided into three steps. 
In step 1, we set the question about the existence of a certain object as the root node. In step 2, we set all questions related to the object of the root node, such as those about its attributes and relations, as corresponding leaf nodes. Finally, in step 3, determine whether the root node question is answered correctly; if not, there is no need to judge the questions at the leaf nodes, and we directly determine that all questions on the tree are answered incorrectly. For instance, after step 1 and step 2, we obtain a list of questions such as $L^Q = \{ Q_1:Independent, Q_2:Depends\ on\ Q_1\}$. Before determining if the answer to $Q_2$ is correct, we first assess $Q_1$, which concerns the accuracy related to the root node. If the question about the existence of an object, which is at the root node, is answered incorrectly, we consider that all other related questions must be hallucinatory.

\section{Experiments}

\begin{table}[t]
  \caption{The number of questions generated by FIHA from various datasets. }
  \label{diff_dataset_qa_result_sum}
  \centering
  \small
  \begin{tabular}{cccc} 
    \toprule
  Source&From Image &From Caption \\ \midrule
    MSCOCO  & 25,699 & 13,007  \\
    Foggy   & 7,232 & 2,801   \\
    Visual Genome  &1,566 &476 \\
    \bottomrule
  \end{tabular}
\end{table}

\begin{table*}
  \caption{
Evaluation results of LVLMs on questions generated from images and captions using FIHA. The upper part is from the MSCOCO dataset and the bottom part is from the Foggy dataset. F1 (Gen) is the BERTScore value calculated from the standard answers and model outputs for all Wh-Questions. For more details, please refer to the explanation of Metrics in Section 3.1.}
  \label{overall_result}
  \centering
\resizebox{0.75\textwidth}{!}{ 

      \begin{tabular}{p{5cm}cccccccccc} 
        \toprule
      \multirow{3}{*}{Model} &  \multicolumn{5}{c}{Question Generated from Image} & \multicolumn{5}{c}{Question Generated from Caption} \\ 
      \cmidrule(r){2-6}\cmidrule(r){7-11}
         &Acc. &P. &R. &F1 &F1 (Gen)
     &Acc. &P. &R. &F1 &F1 (Gen)\\
        \midrule
        \rowcolor{lightgray}{\textit{MSCOCO }} & & & & & & & & & & \\
        mPLUG-Owl & 42.1 & 70.2  & 61.4 & 43.7 & 15.2 & 31.4 & 61.6 & 55.5 & 31.2 &11.4\\
        MiniGPT-4 & 23.5 & 27.5  & 22.2 & 22.1 &21.6 & 15.9 & 25.7 & 28.8 & 14.2 &\cellcolor{gray!15}18.4 \\
        MultiModal-GPT & 59.1 & 46.4  & 47.1 & 46.6 & 16.1 & 23.8 & 39.6  & 45.7 & 22.1 &10.8\\
        LLaVA-1.5-7B& 77.8 & 77.0  & 65.9 & 67.7 & 21.4 & 50.7 & 64.9 & 67.5 & 50.5 &13.7\\
        LLaVA-1.5-13B & 78.9 & 80.9  & 66.4 & 68.3 &20.9 & 47.6 & 64.2 & 65.5 & 48.5 &13.8\\
        InstructBLIP  & \cellcolor{gray!15}84.7 & \cellcolor{gray!30}83.3  & \cellcolor{gray!15}78.6 & \cellcolor{gray!15}80.4 
        & \cellcolor{gray!15}21.8
        &\cellcolor{gray!15} 65.7 & \cellcolor{gray!15}69.5  & \cellcolor{gray!30}77.4 & \cellcolor{gray!15}64.2 
        & 14.1\\
        GPT-4V& \cellcolor{gray!30}87.2 & \cellcolor{gray!15}81.4  & \cellcolor{gray!30}86.3 & \cellcolor{gray!30}85.5 & \cellcolor{gray!30}25.2 & \cellcolor{gray!30}70.3 & \cellcolor{gray!30}71.5 & \cellcolor{gray!15}75.8 & \cellcolor{gray!30}69.3 &
        \cellcolor{gray!30}22.7\\
        \midrule
        
        \rowcolor{lightgray}{\textit{Foggy}}  & & & & & & & & & &\\
        mPLUG-Owl   & 64.8 & 60.2  & 51.1 & 42.7 &18.6 & 29.5 & \cellcolor{gray!30}58.9  & 51.6 & 25.6 &\cellcolor{gray!30}29.3\\
        MiniGPT-4  & 30.1 & 30.2  & 27.6 & 28.1 & 9.4 & 23.4 & 34.4  & 37.8 & 23.0 &11.6\\
        MultiModal-GPT  & 50.2 & 48.7  & 46.1 & 45.8 &17.6  & 28.1 & 43.9  & 47.9 & 25.4 &24.5\\
        LLaVA-1.5-7B& 67.7 & 68.4  & 56.2 & 52.9 &\cellcolor{gray!15}19.7 & 29.1 & 50.0  & 49.2 & 25.8 &27.5\\
        LLaVA-1.5-13B & 68.1 & \cellcolor{gray!15}71.5  & 56.1 & 52.3 &18.8 & 28.9 & 49.2  & 49.8 & 25.5 &27.7\\
        InstructBLIP & \cellcolor{gray!15}70.9 & \cellcolor{gray!30}75.6  & \cellcolor{gray!15}60.2 & \cellcolor{gray!15}58.8 &
        \cellcolor{gray!30}20.3
        & \cellcolor{gray!15}32.8 & \cellcolor{gray!30}58.3  & \cellcolor{gray!30}53.2 & \cellcolor{gray!15}30.5 &
        \cellcolor{gray!15}29.2
        \\
        GPT-4V  & \cellcolor{gray!30}76.3 & 70.1  & \cellcolor{gray!30}64.6 & \cellcolor{gray!30}66.0 &
         16.2
        & \cellcolor{gray!30}33.7 & 53.3 & \cellcolor{gray!15}51.7 & \cellcolor{gray!30}32.1 &
        21.7
        \\
        \midrule
        \rowcolor{lightgray}{\textit{Visual Genome  }} & & & & & & & &  & &\\
        mPLUG-Owl & 41.8 & 68.9  & 60.9 & 43.3 &16.4 & 44.6 & 71.7  & 51.8 & 33.8 &24.1\\
        MiniGPT-4 & 22.9 & 26.8  & 22.0 & 21.8 &22.3 & 15.9 & 25.7 & 28.8 & 14.2 &16.8\\
        MultiModal-GPT & 58.8 & 46.1  & 46.9 & 46.3 &17.2 & 65.3 & 65.2  & 62.2 & 61.7 &21.7\\
        LLaVA-1.5-7B& 77.7 & 77.2  & 61.1 & 67.9 & 20.6 & 56.4 & 73.8  & 62.0 & 52.6 &20.1\\
        LLaVA-1.5-13B & 79.0 & 81.2  & 66.7 & 68.6 & 20.9 & 74.3 & 79.6  & 68.8 & 69.2 &20.2\\
        InstructBLIP  & \cellcolor{gray!15}84.5 & \cellcolor{gray!30}83.7  & \cellcolor{gray!15}79.0 & \cellcolor{gray!15}80.7 &
        \cellcolor{gray!15}22.4 &
        \cellcolor{gray!15}67.7 & \cellcolor{gray!15}78.4  & \cellcolor{gray!15}71.9 & \cellcolor{gray!15}66.7 &
        \cellcolor{gray!30}26.8\\
        GPT-4V& \cellcolor{gray!30}87.0 & \cellcolor{gray!15}81.2  & \cellcolor{gray!30}86.0 & \cellcolor{gray!30}85.3 & 
        \cellcolor{gray!30}23.7 &
        \cellcolor{gray!30}84.2 & \cellcolor{gray!30}78.9 & \cellcolor{gray!30}84.1 & \cellcolor{gray!30}82.2 &
        \cellcolor{gray!15}26.0\\
        \bottomrule
      \end{tabular}
    }
\end{table*}

\subsection{Setup}
\label{Experimental Setup}

\noindent
\textbf{Datasets.} We construct a hallucination evaluation benchmark FIHA-v1 based on three datasets: the MSCOCO \cite{mscoco}, the Foggy \cite{Cordts_2016_CVPR} and Visual Genome \cite{krishna2017visual}. 
\textbf{MSCOCO} is a large image dataset by Microsoft with over 330,000 images. More than 200,000 are annotated across 80 object categories. 
\textbf{Foggy} is a synthetic fog dataset with 1,500 images, each in three fog levels (no fog, medium fog, dense fog).
\textbf{Visual Genome} has 108,077 images with some overlap with MSCOCO.
For our benchmark, we only use the test sets of these datasets to avoid overlap with training data used by LVLMs.

\textbf{Metrics.} We use Accuracy (Acc.), Precision (P.), Recall (R.), and F1 Score (F1) as evaluation metrics for Yes-No questions. For Wh-Questions, we use the F1 (Gen) from BERTScore \cite{DBLP:conf/iclr/ZhangKWWA20} for evaluation. 

\textbf{Models.} We select seven mainstream LVLMs for evaluation: mPLUG-Owl \cite{mPLUG-Owl}, MiniGPT-4 \cite{minigpt4}, MultiModal-GPT \cite{MultiModal-GPT}, LLaVA-1.5-7B \cite{liu2023llava}, LLaVA-1.5-13B \cite{liu2023llava}, InstructBLIP \cite{InstructBLIP}, and GPT-4V \cite{openai2024gpt4}.

\subsection{Data Processing and Analysis}
We randomly selected 500 images from the MSCOCO dataset, 150 images from the Foggy, and 50 from the Visual Genome dataset. Using the process described in Section \ref{sec:method}, we generate tens of thousands of Q\&A pairs. The detailed data statistics of our FIHA-v1 benchmark can be found in Table \ref{diff_dataset_qa_result_sum}.

\begin{figure}[h]
  \centering
  \includegraphics[width=1\linewidth]{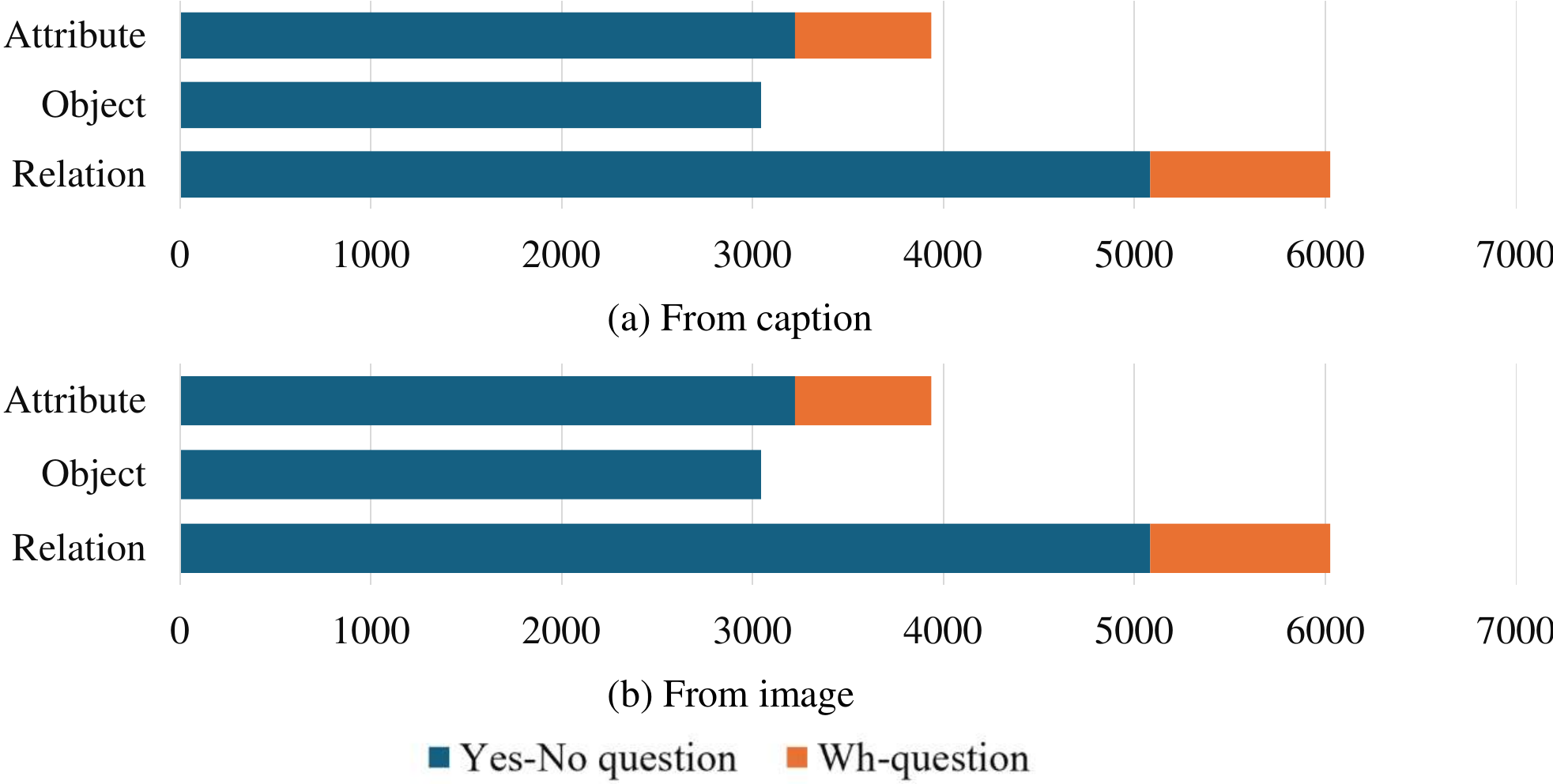}
  \caption{Distribution of two types of question, \ie Yes-No and Wh-questions.}
  \label{distri_image_caption}
\end{figure}

\par
Figure \ref{distri_image_caption} illustrates the distribution of question types generated from images and captions. 
The proportion of questions related to objects, attributes, and relations is relatively balanced, reflecting the rationality of the method design. 
It is noteworthy that the abundance of the question category reflects FIHA's effective capability in generating tasks of the generation type, thereby enabling a more effective assessment of hallucinations.

\begin{table*}
  \caption{The fine-grained assessment of LVLMs evaluates object, attribute, and relation accuracy using Q\&A pairs generated from captions in the MSCOCO and Foggy Cityscapes datasets.}
  \label{fine-grained-image}
  \centering
  
  \resizebox{0.8\textwidth}{!}{
      \begin{tabular}{lcccccccccccccc} 
        \toprule
      \multirow{3}{*}{Model} &  \multicolumn{4}{c}{Object} & \multicolumn{4}{c}{Attribute} & \multicolumn{4}{c}{Relation}\\ 
      \cmidrule(r){2-5}\cmidrule(r){6-9}\cmidrule(r){10-13}
      &Acc.   &P. &R. &F1  &Acc. &P. &R. &F1 &Acc.
     &P. &R. &F1 \\
        \midrule
        \rowcolor{lightgray}{\textit{MSCOCO}}  & & & & & & & & & & & &\\
        mPLUG-Owl  & 57.3 & 75.7  & 47.3 & 48.0 & 20.6 & 55.7 & 53.5 & 20.4 & 22.7 & 56.5 & 55.8 & 22.7\\
        MiniGPT-4  & 66.2 & 59.5  & 62.6 & 59.5 & 9.6 & 12.8 & 9.2 & 9.4 & 4.7 & 12.1 & 11.4 & 4.9\\
        MultiModal-GPT  & 51.6 & 54.1  & 51.5 & 42.5 & 16.0 & 39.2 & 42.8 & 15.8 & 12.1 & 30.8 & 39.6 & 11.8\\
        LLaVA-1.5-7B & 79.2 & 82.4  & 77.5 & 78.4 & 27.9 & 55.6 & 56.7 & 27.8 & 47.9 & 59.1 & 69.7 & 44.7 \\
        LLaVA-1.5-13B  & 70.8 & 80.6  & 70.2 & 68.3 & 34.3 & 56.4 & 59.7 & 33.7 & 42.1 & 58.3 & 66.6 & 48.1\\
        InstructBLIP  & \cellcolor{gray!15}84.6 & \cellcolor{gray!30}87.7  & \cellcolor{gray!15}81.4 & \cellcolor{gray!15}84.2 & \cellcolor{gray!15}61.0 & \cellcolor{gray!15}62.2 & \cellcolor{gray!15}76.2 &\cellcolor{gray!15}55.6 & \cellcolor{gray!15}57.5 & \cellcolor{gray!15}61.0 & \cellcolor{gray!30}75.7 & \cellcolor{gray!15}52.1\\
        GPT-4V   & \cellcolor{gray!30}90.8 & \cellcolor{gray!30}87.7  & \cellcolor{gray!30}89.8 & \cellcolor{gray!30}88.6 & \cellcolor{gray!30}83.6 & \cellcolor{gray!30}77.7 & \cellcolor{gray!30}85.2 & \cellcolor{gray!30}79.8 & \cellcolor{gray!30}66.2 & \cellcolor{gray!30}61.2 & \cellcolor{gray!15}73.2 & \cellcolor{gray!30}58.3\\
        \midrule
        \rowcolor{lightgray}{\textit{Foggy}}  & & & & & & & & & & & &\\
        mPLUG-Owl  & 52.9 & 32.3  & 50.0 & 39.2 & \cellcolor{gray!15}15.7 & \cellcolor{gray!15}54.8 & \cellcolor{gray!15}52.1 & \cellcolor{gray!15}15.3 & 11.8 & 34.6 & 46.9 & 11.1\\
        MiniGPT-4  & \cellcolor{gray!30}62.1 & 60.6  & \cellcolor{gray!15}58.4 & \cellcolor{gray!30}57.8 & 9.6 & 25.1 & 14.6 & 9.3 & 8.5 & 23.2 & 26.5 & 8.5\\
        MultiModal-GPT & 52.9 & 59.7  & 52.6 & 42.1 & 12.6 & 33.9 & 38.6 & 12.5 & 11.5 & 33.3 & 39.4 & 11.4 \\
        LLaVA-1.5-7B & 54.0 & 63.3  & 54.0 & 44.4 & 11.5 & 33.2 & 46.0 & 10.8 & 15.4 & 47.8 & 48.9 & 15.1 \\
        LLaVA-1.5-13B  & 54.2 & 62.8  & 54.2 & 44.6 & 11.3 & 31.4 & 46.3 & 10.6 & 14.9 & 47.0 & 48.6 & 14.6\\
        InstructBLIP  & 54.2 & \cellcolor{gray!15}65.2  & 53.9 & 44.2 & \cellcolor{gray!30}20.7 & \cellcolor{gray!30}55.1 & \cellcolor{gray!30}54.6 & \cellcolor{gray!30}20.6 & \cellcolor{gray!15}15.9 & \cellcolor{gray!15}48.5 & \cellcolor{gray!15}49.2 & \cellcolor{gray!15}15.6\\
        GPT-4V  & \cellcolor{gray!15}61.8 & \cellcolor{gray!30}69.6  & \cellcolor{gray!30}59.2 & \cellcolor{gray!15}54.5 & 11.1 & 37.0 & 33.1 & 11.0 & \cellcolor{gray!30}20.4 & \cellcolor{gray!30}50.5 & \cellcolor{gray!30}50.4 & \cellcolor{gray!30}20.3\\
        \bottomrule
      \end{tabular}
    }
\end{table*}

\subsection{Experimental Results}
\subsubsection{Overall Results on Datasets Generated by FIHA}
\label{Overall Results}
We show the hallucination comparison of the seven mainstream LVLMs on our FIHA-v1 in Table \ref{overall_result}. From this Table, we have several observations.
1) It's worth highlighting that GPT-4V excels in both image and caption Q\&A pairs, achieving the best performance among the evaluated models. 
2) The second-best performer is InstructBLIP, which significantly outperforms other models except GPT-4V across most metrics. 
3) Additionally, we have observed that model parameters are also significant factors affecting performance. For instance, LLaVA-1.5-13B provides a more comprehensive improvement over the LLaVA-1.5-7B. 

In addition, we also show the performance of 7 mainstream LVLMs on FIHA-v1 based on the Visual Genome dataset. 
The results show a similar trend as compared to the performance in MSCOCO datasets. 
Specifically, the GPT-4V performs best and MiniGPT-4 performs the worst. LLaVA-1.5-13B performs better than LLaVA-1.5-7B, which also indicates that the model parameter size influences the performance.

\subsubsection{Fine-Grained Results}
Furthermore, we evaluate the model's performance from more dimensions (\ie the object existence, attribute, and relation) with FIHA. 
We show the fine-grained evaluation results in Table \ref{fine-grained-image}.

\paragraph{Object Hallucination} 
From the results for the object, we can observe that even after introducing more negative samples, the \textit{Accuracy} and \textit{Precision} of the models remain high, indicating that most models have a strong capability to determine whether an object exists or not. 
In comparison, the \textit{Recall} is somewhat lower, indicating that the model still has a tendency to lean towards affirmative responses.

\paragraph{Attribute Hallucination} 
From the results for the attribute, tt is evident that this part of the hallucination is much more difficult to identify. Compared to the object itself, its color, quantity, size, and so on are indeed more challenging to judge. Even the best-performing GPT-4V has an F1 score of less than 80 on regular data. Moreover, the performance of the vast majority of models plummets on special datasets, indicating that the robustness of existing LVLMs needs to be enhanced.

\paragraph{Relation Hallucination} 
From the results for the relation, this part is the most challenging, with the F1 score of GPT-4V on regular data not even reaching 60\%. The potential reason is that Q\&A pairs of the relation types involved more than one object, which makes it challenging.

\section{Analysis}
In this section, we further evaluate the effectiveness of our benchmark FIHA-v1 by four research questions.
    
\subsection{How Reliable is FIHA?}
\label{44}

To evaluate the benchmark's reliability, we manually check the accuracy of Q\&A pairs in FIHA-v1 generated by the pipeline, verifying if the answers match the questions. In the human evaluation process, we employ annotators manually evaluate whether the answer is right in each image, question, and answer pair. Whether it is a Yes-No question or a Wh-Question, it will be marked as True or False, that is, whether the answer is right for the question. The final accuracy is calculated as (num\_true / (num\_true + num\_false)).
Table \ref{diff_dataset_qa_result} shows that Q\&A pairs from image captions are 96\% accurate on MSCOCO samples.
The MSCOCO-based pipeline using Grounding DINO achieves 98.2\% accuracy, with 1.8\% errors, such as missing details or misidentifying colors.
Overall, FIHA shows high reliability in generating datasets for evaluating hallucinations in LVLMs, with near-perfect accuracy in caption-based datasets.

\begin{table}[t]
  \caption{
The results of human evaluation (accuracy) of Q\&A pairs generated from different datasets.}
  \label{diff_dataset_qa_result}
  \centering
  \small
  \begin{tabular}{ccc} 
    \toprule
  Source&From Image &From Caption \\
    \midrule
    MSCOCO   & 98.2 & 96.0  \\
    No Foggy  & 98.1 & 96.1   \\
    Medium Foggy  & 97.6 & 94.5 \\
    Dense Foggy & 96.3 & 94.1  \\
    \bottomrule
  \end{tabular}
  \vspace{-3mm}
\end{table}

We also test on complex images using the Foggy dataset \cite{Cordts_2016_CVPR}, which test evaluates noise's effect on the framework's accuracy. Examples are in Appendix \ref{example_foggy}. 

As shown in Table \ref{diff_dataset_qa_result}, under dense fog, the accuracy for \textit{Q\&A pairs generation from images} is 96.3\%, while for \textit{Q\&A pairs generation from captions}, it is 94.1\%. For medium fog, the accuracies are 97.6\% and 94.5\%, respectively. Under no fog, the accuracies are 98.1\% and 94.1\%, similar to MSCOCO results.

The results show that as fog increases, the accuracy of FIHA's Q\&A pairs decreases, highlighting the challenge of blurry images. 



\subsection{What is the Impact of Introducing DSG?}


To improve hallucination assessment, we propose the DSG mechanism, which models dependencies between hallucination types. Table \ref{tab:decrease} shows that stronger models like GPT-4V and LLaVA-1.5-13B exhibit smaller performance drops (6.0\% and 2.7\%), indicating their robustness to dependency-based evaluations. In contrast, weaker models such as MultiModal-GPT and mPLUG-Owl show substantial declines (21.3\% and 29.6\%), reflecting frequent root-level errors that propagate to leaf-level questions, which these models might otherwise answer correctly under standard evaluation.

The results reveal that most hallucination errors occur at fundamental levels, such as object recognition, with models like LLaVA-1.5-13B maintaining high accuracy due to fewer cascading errors. Precision drops more than recall across models, suggesting that DSG effectively uncovers false positives. For example, MiniGPT-4’s precision decreased by 11.0\%, highlighting previously unnoticed errors.

While InstructBLIP shows a significant F1 drop (from 9.6\% to 5.7\%), GPT-4V remains relatively stable (9.9\% to 8.4\%), suggesting stronger contextual reasoning. These results demonstrate that DSG provides a rigorous evaluation, exposing model weaknesses that standard assessments may miss.

\begin{table}[t]
  \caption{The performance decrease of various LVLMs after introducing DSG.}
  \label{DSG_result}
  \centering
  \resizebox{0.8\columnwidth}{!}{
    \begin{tabular}{lcccccc}
      \toprule
      Model & Acc.$\downarrow$ & P.$\downarrow$ & R.$\downarrow$ & F1$\downarrow$ & F1 (Gen)$\downarrow$\\
      \midrule
      mPLUG-Owl  & 29.6 & 22.1 & 14.0 & 28.7 &14.2\\
      MiniGPT-4  & 62.6 & 51.8 & 62.1 & 61.2 &42.3\\
      MultiModal-GPT  & 21.3 & 27.6 & 21.9 & 24.3 &12.9 \\
      LLaVA-1.5-7B  & \cellcolor{gray!15}4.2 & 11.7 & \cellcolor{gray!15}4.5 & \cellcolor{gray!15}4.8 & 5.7\\
      LLaVA-1.5-13B  & \cellcolor{gray!30}2.7 & \cellcolor{gray!30}8.1 & \cellcolor{gray!30}3.3 & \cellcolor{gray!30}3.6  &\cellcolor{gray!15}5.1\\
      InstructBLIP  & 5.7 & \cellcolor{gray!15}9.6 & 5.7 & 5.7  &6.9\\
      GPT-4V & 6.0 & 9.9 & 5.4 & 8.4 & \cellcolor{gray!30}3.9 \\
      \bottomrule
    \end{tabular}
  }
  \label{tab:decrease}
  \vspace{-2mm}
\end{table}

\subsection{How and Why LLM is Free?}
FIHA has a big benefit: it doesn't need extra big language models like GPT-4. This is shown when we make questions from true information. We do this with Python code, and you can find more details in Section \ref{Question Answer Pair Generation} and Appendix \ref{code Experimental Setup}. 

We don't use big language models to help make questions because they cost a lot. For example, if someone wants to make questions for 500 pictures, they would need about 36,900 questions (like we saw with MS COCO). This would cost almost \$400 in API fees, which is very expensive.

\subsection{Is the Information Extracted from Images More Comprehensive?}
\begin{figure}[h]
  \centering
  \includegraphics[width=\linewidth]{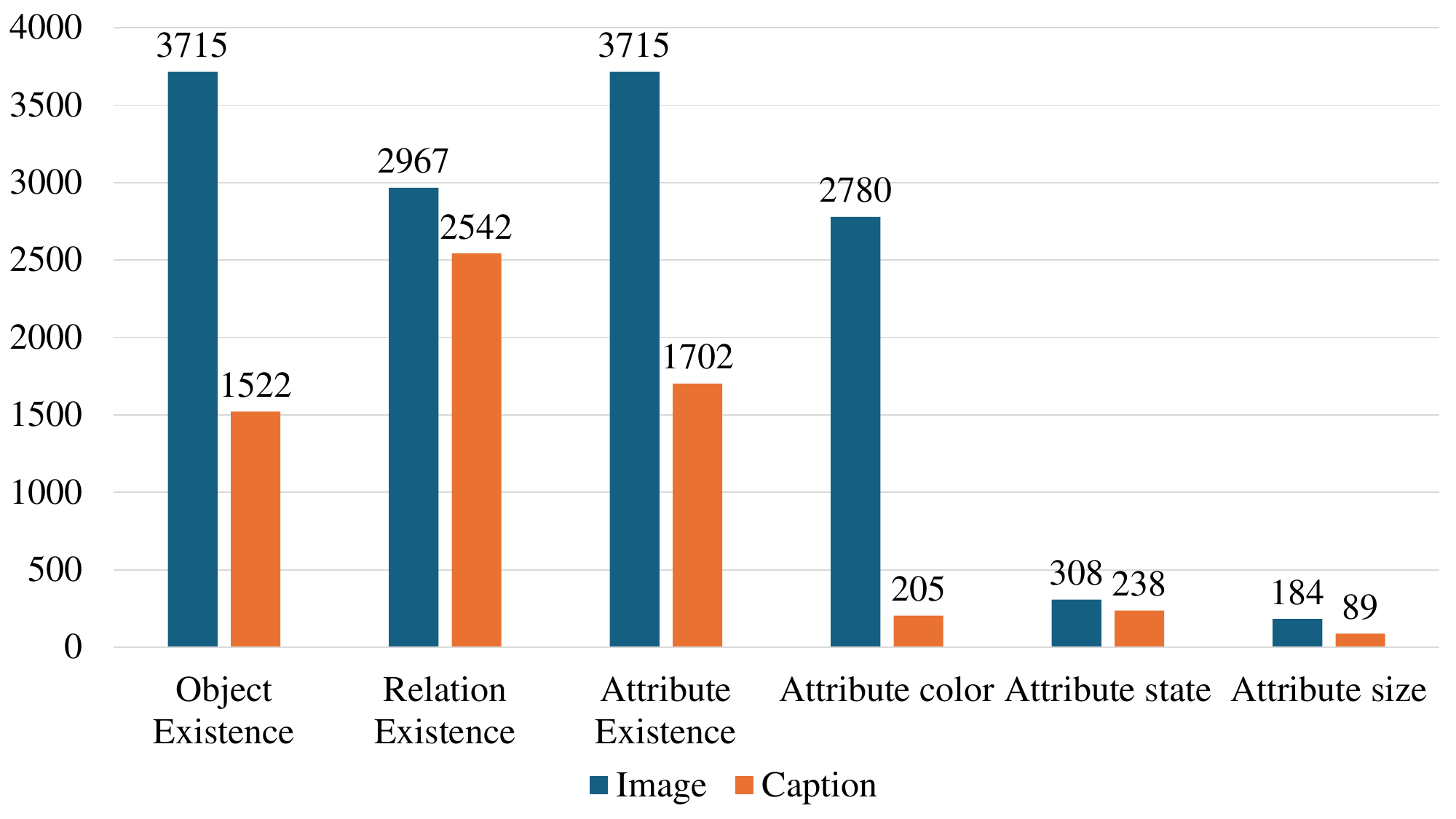}
  \vspace{-6mm}
  \caption{Comparison of the number of Q\&A paris across different types of hallucination from image and caption.}
\vspace{-6mm}
  \label{diff_img_cap}
\end{figure}

As shown in Figure \ref{block}, we extract information from both the image and the caption to construct Q\&A pairs. Typically, the image itself contains more abundant information. In this section, we will verify whether the information extracted from the image is more comprehensive and diverse than that extracted from the caption.
We have separately counted the number of six different types of Q\&A pairs from image and caption, mainly focusing on the three directions of object, attribute, and relation. 
As shown in Figure \ref{diff_img_cap}, it is evident that the information extracted from the image surpasses the information extracted from the caption. 

\subsection{Why are Performance on our Benchmark  Lower Than Others?}
In the experiment, we observe that our test results are lower than others, \eg POPE \cite{li2023evaluating} and HaELM \cite{wang2023evaluation}, indicating that FIHA can detect more difficult and distinct issues. 
We analyze that there are mainly three reasons: 
firstly, we added a large number of misleading negative samples, and since the model tends to give affirmative answers (Section \ref{Overall Results}), this increases the difficulty of evaluation. 
Secondly, the role of DSG directly impacts the results and improves the reliability of the evaluation method. 
Finally, the comprehensiveness of FIHA is more challenging than methods that focus primarily on generating coarse-grained object-level questions.

\subsection{Will Using Fixed Templates Limit the Diversity and Types of Questions?}
According to the description of LVLM hallucinations in the existing work \cite{bai2024hallucination}, the types of questions usually limited, even if the questions are generated by LLMs. We compared the method \cite{chen2024unified,jiang2024halevaluniversalfinegrainedhallucination} of using LLMs to generate questions and found that the diversity of questions generated by the LLM-based method is similar to our method.

\subsection{Why Are the Metrics Lower on the Foggy Dataset Compared to MSCOCO?}
There are mainly two reasons: 1. The Foggy dataset is a collection of images captured in foggy weather conditions, which inherently falls under the category of complex scenes (such as nighttime, underwater, rainy conditions, etc.). Due to reduced visibility in foggy environments, many objects in the images become blurred, increasing the difficulty for models to accurately recognize them. 2. The training set of MSCOCO is commonly used for training various LVLMs, while datasets of the Foggy type are rarely used for training. Therefore, we can see Foggy as a out-of-distribution test setting. This results in the various LVLMs being unfamiliar with the Foggy style.

\section{Related Work}

In this section, we mainly discuss existing Large Vision-Language Models (LVLMs) and the hallucination problems that exist in LVLMs.

\paragraph{Large Vision-Language Model}
With the success of pretraining in LLMs \cite{touvron2023llama} and VFMs \cite{DBLP:journals/corr/abs-2307-13721}, many researchers \cite{alayrac2022flamingo,li2023otter} extended LLMs to understand real-world images through LVLMs, benefiting from in-context and few-shot learning. This led to a rise in visual instruction-adapted LVLMs \cite{liu2023llava,minigpt4,InstructBLIP,MultiModal-GPT}, which show strong generalization across VL tasks. Most use GPT-4 to generate multimodal instruction datasets and multi-stage pretraining to align visual data with LLMs. For example, Liu \etal \cite{liu2023llava} aligned LLaMA \cite{touvron2023llama} with a visual encoder output, while Zhu \etal \cite{minigpt4} fine-tuned Vicuna \cite{peng2023instruction} for cross-modal alignment. Similarly, Multimodal GPT \cite{MultiModal-GPT} and InstructBLIP \cite{InstructBLIP} used VL datasets, with the former using BLIP2 \cite{li2023blip} and the latter starting from Flamingo \cite{alayrac2022flamingo}.  

Despite these advances, LVLMs still struggle with hallucinations in textual output, limiting their effectiveness in vision-language tasks \cite{rohrbach2018object}.

\paragraph{Hallucination in LVLMs} 
Recent studies have focused on hallucination in LVLMs. Some works, summarized in Table 1, address hallucination detection and evaluation \cite{li2023evaluating,wang2023evaluation,wang2023llm,jing2023faithscore}, while others propose mitigation methods \cite{liu2023mitigating,zhou2023analyzing,yin2023woodpecker,jing2024fgaif,jing2025comprehensiveanalysisvisualobject}. For instance, Bingo \cite{Bingo} evaluates GPT-4V's hallucinations with bias and interference, and HallusionBench \cite{HallusionBench} diagnoses entangled language hallucinations and visual illusions. AutoHallusion \cite{AutoHallusion} generates benchmarks by manipulating images to challenge language priors, and Hal-Eval \cite{jiang2024halevaluniversalfinegrainedhallucination} categorizes hallucinations into objects, attributes, relations, and events.  

Despite progress, fine-grained detection is less explored. Li \etal \cite{li2023evaluating} introduced POPE to evaluate object-level hallucinations, showing LVLMs’ susceptibility. Wang \etal \cite{wang2023evaluation} proposed HaELM, creating a hallucination dataset and fine-tuning LLaMA for detection. These methods focus on object-level issues or require training. To address limitations, Wang \etal \cite{wang2023llm} developed AMBER, a benchmark for generative and discriminative tasks involving object, attribute, and relation hallucinations, though it depends on human annotations.  

Existing methods mostly primarily use LLMs to extract key information from image captions, such as objects, colors, positions, etc. Based on the extracted keywords and different prompts, the LLM then generates corresponding questions. For each generated question, the LLM also provides the corresponding answer. It can be seen that every step relies entirely on the LLM and manually designed prompts. In contrast, our method does not rely on LLMs for any step, from key information extraction to question and answer generation. When generating questions, it fills in pre-defined, well-structured templates, ensuring both accuracy and controllable question types. Therefore, our method is effeicent and not costly.

\section{Conclusion}
In recent years, large vision-language models have developed quickly, but hallucinations remain a serious concern.
Current hallucination evaluation methods face problems like high costs, limited scope, and lack of generalization. Thus, we introduce FIHA, a multi-dimensional detection method that requires no LLMs and no annotations. FIHA can automatically create high-quality Q\&A pairs for any image dataset. 
We conducted a thorough analysis of the performance of mainstream LVLMs, identified the issues, and proposed potential methods for improvement. In the future, we will delve deeper into methods for alleviating hallucinations.

\section*{Liminations}
FIHA has comprehensive features and maintains a high overall quality. Despite the limitations discussed in the previous analysis section, there are additional constraints in some aspects. 
The generated Q\&A primarily focuses on the existence, attributes, and relations of main objects in the images, while lacking in Q\&A for surrounding and minor objects. This is due to the FRCNN's lower confidence in detecting small and less obvious objects. 




\bibliography{custom}
\bibliographystyle{acl_natbib}

\newpage
\appendix



\clearpage

\section{Code Example for Generating QA Pairs Based on Extracted Information}
\label{code Experimental Setup}

\lstset{
  language=Python,              
  basicstyle=\ttfamily\footnotesize,   
  keywordstyle=\color{blue},    
  stringstyle=\color{red},      
  commentstyle=\color{green},   
  showspaces=false,             
  showtabs=false,               
  frame=single,                 
  numbers=left,                 
  numberstyle=\small\color{gray},
  breaklines=true,              
  tabsize=4,                    
}

\begin{lstlisting}
if relation.endswith(tuple(['ing', 'ed'])):
    question = f"Is the {subject} {relation} the {object} in the image?"
elif relation.endswith(tuple(['over', 'under', 'above', 'near', 'behind', 'on', 'at'])):
    if obj_is_living(object):
        question = f"Who is {relation} the {object} in the image?"
    else:
        question = f"What is {relation} the {subject} in the image?"
\end{lstlisting}

\section{Example of foggy Cityscapes Images datasets }
\label{example_foggy}
\begin{figure}[h]
  \centering
  \includegraphics[width=\linewidth]{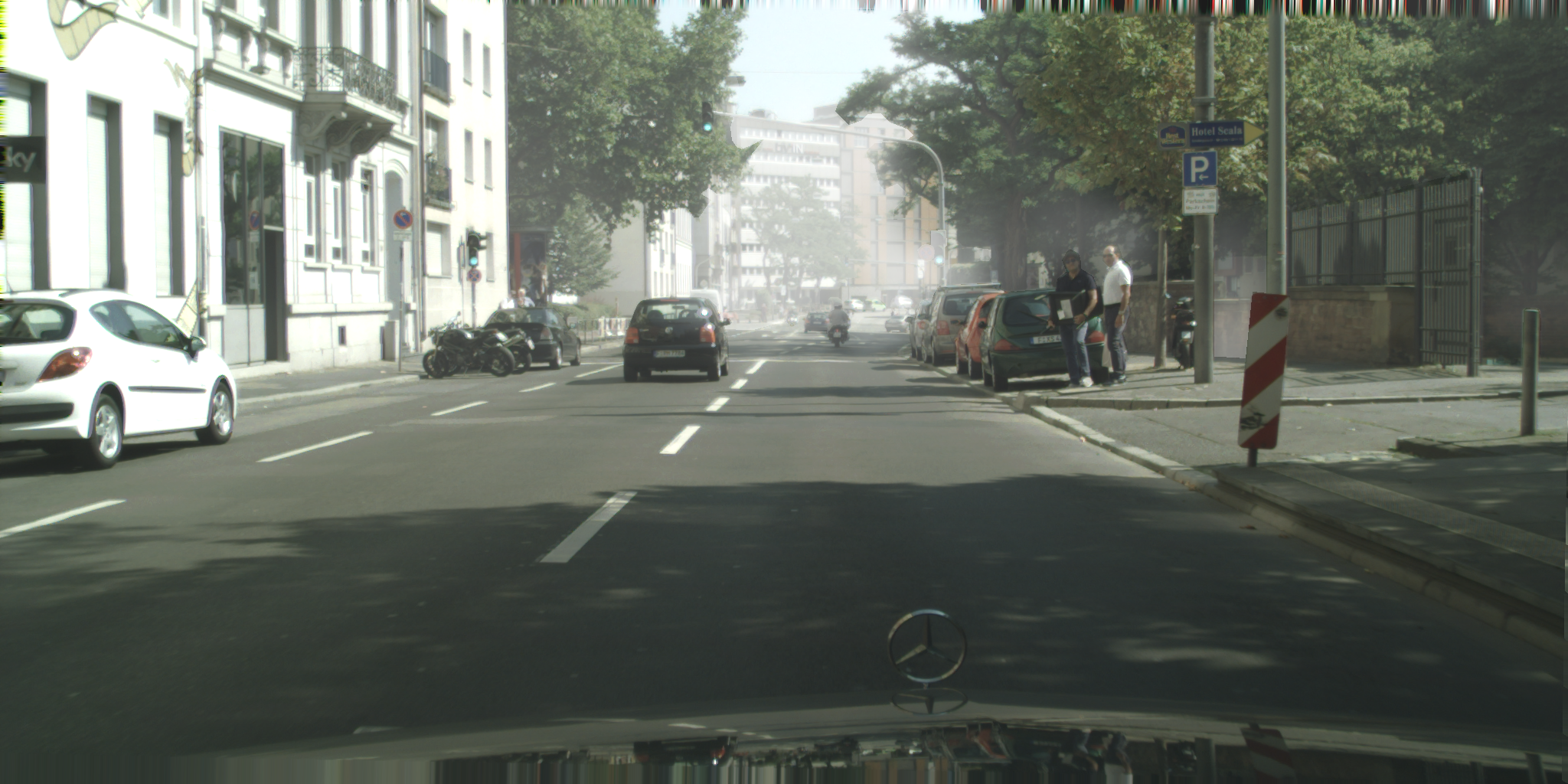}
\caption{no foggy}
    \includegraphics[width=\linewidth]{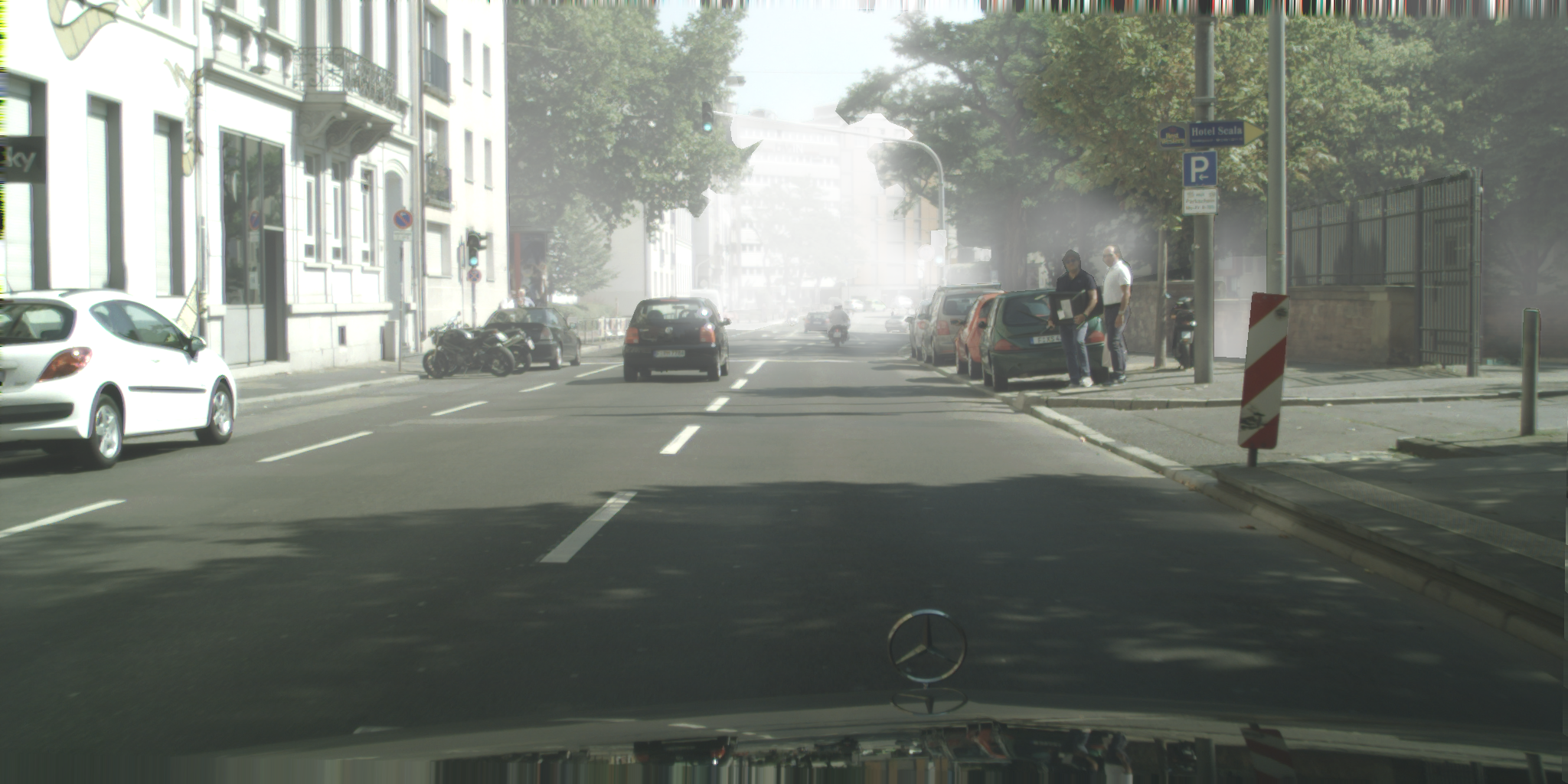}
    \caption{medium foggy}

    \includegraphics[width=\linewidth]{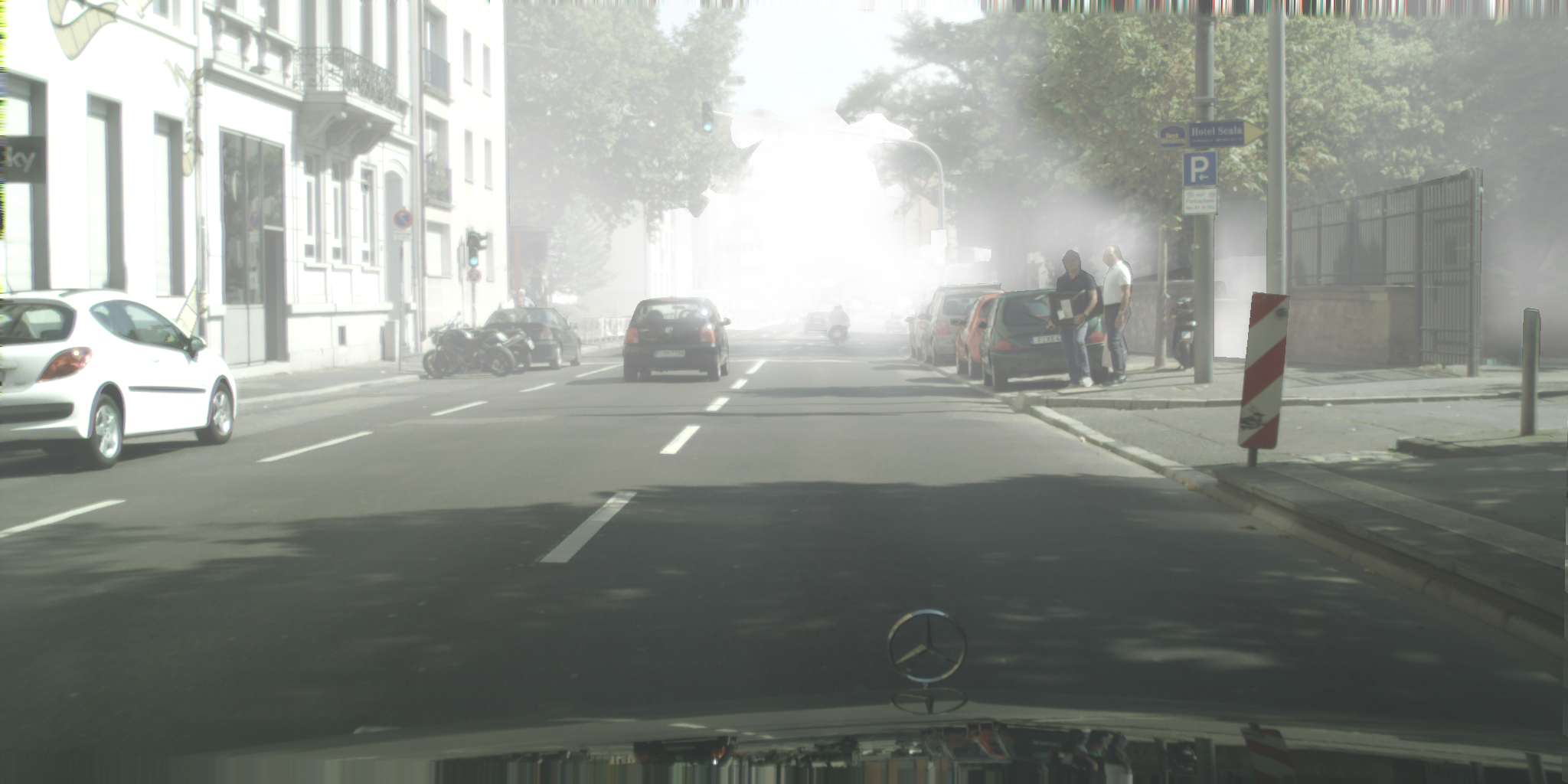}
    \caption{dense foggy}

\end{figure}

\section{Expalnation for DSG}
\label{app:explanation}
\begin{figure*}[h]
  \centering
  \includegraphics[width=1\linewidth]{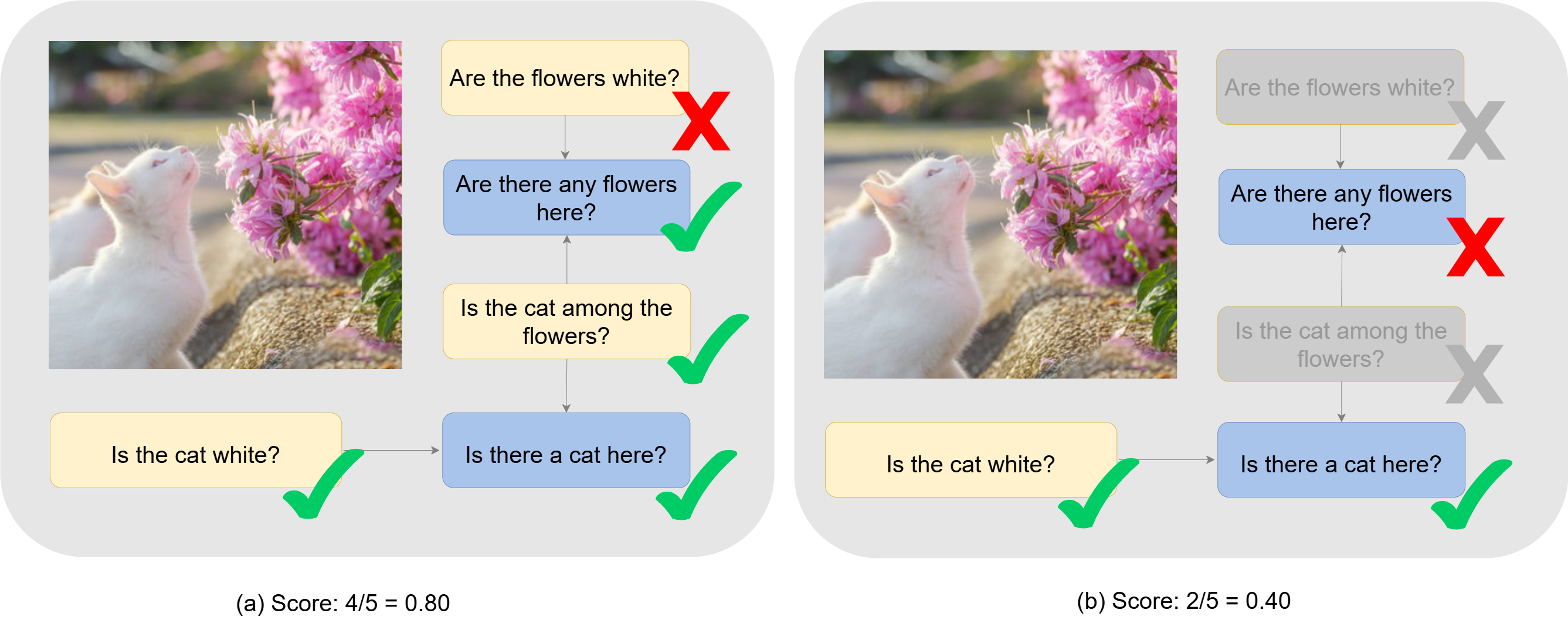}
  \caption{The diagram on the right is a schematic illustration of the impact on the results after the introduction of DSG.}
  \label{dsg_figure}
\end{figure*}

Depending on the answers, some questions in the hallucination benchmark become invalid and thus should not be asked to the LVLM to evaluate hallucination.  
As shown in Figure \ref{dsg_figure}, if the answer to “are there any flowers here?” is no, dependent questions like “are the flowers white?” should be skipped – 
the LVLM may often say “the flower doesn't exist, but it is white”.

\end{document}